\documentclass[5p,authoryear]{elsarticle}
\myfooter[R]{\today}

\usepackage{fontspec}
\usepackage{amsmath,amssymb,amsthm}
\usepackage{booktabs}
\usepackage{microtype}
\biboptions{authoryear,round}
\usepackage[hidelinks]{hyperref}
\newcommand{\R}{\mathbb R}
\newcommand{\Ndim}{\operatorname{Ndim}}
\newcommand{\conv}{\operatorname{conv}}
\newcommand{\Span}{\operatorname{span}}
\newcommand{\rank}{\operatorname{rank}}
\newcommand{\sgnset}{\{-1,+1\}}
\newcommand{\one}{\mathbf 1}
\DeclareMathOperator*{\argmax}{arg\,max}

\newtheorem{theorem}{Theorem}
\newtheorem{proposition}{Proposition}
\newtheorem{lemma}{Lemma}
\theoremstyle{definition}
\newtheorem{definition}{Definition}
\newtheorem{example}{Example}

\theoremstyle{remark}

\begin{document}
\begin{frontmatter}
\title{Exact Natarajan Dimension of Homogeneous Multiclass Linear Predictors}

\author[aff1]{Jiahua Liu}
\ead{13911879260@163.com}

\author[aff1]{Benchong Li\corref{cor1}}
\ead{libenchong@xidian.edu.cn}

\cortext[cor1]{Corresponding author.}
\affiliation[aff1]{
  organization={School of Mathematics and Statistics, Xidian University},
  city={Xi'an},
  country={China}
}

\begin{abstract}
The Natarajan dimension is a fundamental measure of the capacity of
multiclass hypothesis classes and plays an important role in analyzing
their generalization ability. In this paper, by establishing lower and
upper bounds, we show that the exact
Natarajan dimension of homogeneous multiclass linear predictors with
pairwise distinct class weights is \(n(k-1)-1\), where $n$ is the dimension
of the feature space and $k$ is the number of classes,
for \(n,k\ge3\), \(n,k\in\mathbb N\).
\end{abstract}
\begin{keyword}
Natarajan dimension \sep Multiclass classification
\sep Homogeneous linear predictors \sep Shattering
\end{keyword}
\end{frontmatter}

\section{Introduction}\label{sec:introduction}

In this paper, we consider multiclass classification. In this setting,
a predictor assigns each feature vector a label from a finite set.
Understanding the capacity of multiclass hypothesis classes is a
basic question in learning theory. For binary classes, this capacity
is described by the Vapnik--Chervonenkis
dimension~\citep{vapnik1971,blumer1989}. To extend this notion to
multiclass prediction, \citet{natarajan1989} introduced the Natarajan
dimension, defined via shattering with two distinct candidate labels
fixed at each point. Its role in multiclass learnability
and empirical risk minimization has been studied in
\citet{bendavid1995,daniely2015}.

\citet{daniely2012} analyzed the sample complexity of several
multiclass hypothesis classes using lower bounds for the Natarajan
dimension and upper bounds for the graph dimension.
\citet{daniely2014} established an upper bound on the
Natarajan dimension of their generalized linear hypothesis class
in terms of the dimension of the class-specific feature representation.
Related bounds are also available for decision trees, random forests,
and neural networks~\citep{jin2023}.

A standard approach associates a linear score with each label and
predicts a label of maximum score~\citep{shalevshwartz2014}.
Let \(n\) denote the dimension
of the feature space and \(k\) the number of classes. For
\(x\in\R^n\) and \(c\in[k]\), where \([k]=\{1,\ldots,k\}\), the block feature map
\(\Psi(x,c)\in\R^{nk}\) places \(x\) in the \(c\)-th
\(n\)-dimensional block and zeros in all other blocks.
Predictors in the homogeneous multi-vector construction
use scores \(W\cdot\Psi(x,c)=w_c\cdot x\), with arbitrary
\(W=(w_1,\ldots,w_k)\in\R^{nk}\), and select the smallest
maximizing label.
For the unrestricted multi-vector class,
\citet{shalevshwartz2014} gave the lower bound \((n-1)(k-1)\)
and the upper bound \(nk\) on the Natarajan dimension for \(n,k\ge2\).
In addition, \citet{pan2026} established the improved upper bound
\(n(k-1)\) for the class considered in their paper when \(n,k\ge2\),
and determined its exact dimension to be \(k\) when \(n=2\) and \(k\ge3\).
The bound \(n(k-1)\) equals the number of free parameters after
translating all class weights so that \(w_1=0\). We investigate whether
this bound can be attained when the class weights are pairwise distinct.

In this paper, we denote by \(\mathcal H_\Psi\) the class of homogeneous
multiclass linear predictors with pairwise distinct class weights and
show that its Natarajan dimension is \(n(k-1)-1\) when \(n,k\ge3\).
For the lower bound, we choose class weights as vertices of a polytope
and use its exposed edges to select feature vectors with exactly two
maximizing labels. We select \(n(k-1)-1\) vectors whose comparison rows
are linearly independent, allowing the score differences within their
candidate pairs to be adjusted independently. Sufficiently small weight
perturbations then realize every candidate labeling while preserving
both candidates' strict score advantage over other labels. For the upper bound, we assume that \(n(k-1)\) feature vectors can be
shattered and interpolate the weights realizing their candidate labelings
over a cube. Induction on the dimension of the feature space shows that
the resulting comparison map is nonzero on the cube boundary. The
associated sphere self-map has degree zero because it omits a point,
but degree one because it is homotopic to the identity. This contradiction
establishes the upper bound.

Section~\ref{sec:preliminaries} introduces the necessary preliminaries
and states the main theorem. We then prove the lower and upper bounds
in Sections~\ref{sec:lower} and~\ref{sec:upper}, respectively, and conclude
the paper in Section~\ref{sec:conclusion}.

\section{Preliminaries}\label{sec:preliminaries}

We first recall the definitions of shattering and the Natarajan dimension
and specify the class of homogeneous multiclass linear predictors studied
in this paper. We then introduce the geometric and topological concepts
and results needed for the proofs. The section concludes with the main
theorem.

\subsection{Natarajan dimension}

We follow the notation and definitions in
\citet{shalevshwartz2014}.

\begin{definition}[Shattering]\label{def:shattering}
Let \(\mathcal H\subseteq[k]^{\mathcal X}\).
A finite set \(S\subseteq\mathcal X\) is shattered
by \(\mathcal H\) if there exist fixed functions \(f_0,f_1:S\to[k]\),
with \(f_0(x)\ne f_1(x)\) for every \(x\in S\), such that for every
\(B\subseteq S\), there is \(h_B\in\mathcal H\) satisfying, for every
\(x\in S\),
\[
h_B(x)=
\begin{cases}
f_0(x), & x\in B,\\
f_1(x), & x\in S\setminus B.
\end{cases}
\]
\end{definition}

Using this notion of shattering, we define the Natarajan dimension as follows.

\begin{definition}[Natarajan dimension]\label{def:dimension}
The Natarajan dimension of \(\mathcal H\), denoted by
\(\Ndim(\mathcal H)\), is the maximal size of a set
\(S\subseteq\mathcal X\) shattered by \(\mathcal H\).
If arbitrarily large finite sets are shattered, the dimension is infinite.
\end{definition}

For \(S=\{x_1,\ldots,x_m\}\), order each candidate pair
\(\{f_0(x_i),f_1(x_i)\}\) as \(a_i<b_i\).
Shattering requires every labeling in
\(\{a_1,b_1\}\times\cdots\times\{a_m,b_m\}\) to be realizable.

\subsection{Homogeneous multiclass linear predictors}

We now consider linear scores defined by the standard Euclidean
inner product and use the associated Euclidean norm throughout.
The score-maximization representation is standard 
\citep{daniely2012,shalevshwartz2014}.

\begin{definition}[Homogeneous multiclass linear predictors]\label{def:model}
For \(W=(w_1,\ldots,w_k)\in\R^{nk}\), define
\begin{equation}\label{eq:model}
h_W(x)=\argmax_{c\in[k]}w_c\cdot x,
\qquad x\in\R^n.
\end{equation}
If several labels maximize \(w_c\cdot x\), we choose the smallest one.
The hypothesis class \(\mathcal H_\Psi\) consists of all such \(h_W\)
with pairwise distinct class weights, that is, \(w_c\ne w_d\) for
all \(c,d\in[k]\) with \(c\ne d\).
\end{definition}

Distinct weights can give equal scores at a particular feature vector.
A common translation of all weights preserves score differences, ties,
and pairwise distinctness. We may therefore use effective weights
\begin{equation}\label{eq:effective}
v_1=0,\qquad v_c=w_c-w_1, \qquad c \in \{2,\ldots,k\}.
\end{equation}
For each nonnegative integer \(r\), define the effective
parameter space corresponding to the feature space \(\R^r\) by
\begin{equation}\label{eq:Vr}
V_r=\R^{rq},\qquad \dim V_r=rq,
\end{equation}
where \(q=k-1\). For \(v=(v_2,\ldots,v_k)\in V_r\), each \(v_c\in\R^r\),
\(c \in \{2,\ldots,k\}\). Setting \(v_1=0\) gives the full
tuple \((v_1,\ldots,v_k)\) of class weights. The admissible effective
parameters form
\[
\mathcal A_r=\{v\in V_r:v_c\ne v_d\text{ for all }c,d\in[k],\ c\ne d\},
\qquad v_1=0.
\]

\subsection{Faces of convex polytopes}\label{subsec:polytope-preliminaries}

We introduce some basic concepts from polytope theory that will be used below. For further details, we refer the reader to \citet{ziegler1995}.

A convex polytope is the convex hull of finitely many points.
Specifically, the convex hull of \(u_1,\ldots,u_s\) consists of all convex
combinations \(\sum_{i=1}^s\lambda_i u_i\), where \(\lambda_i\ge0\) and
\(\sum_{i=1}^s\lambda_i=1\). For a nonempty convex polytope \(P\subseteq\R^n\), let
\(A=\operatorname{aff}(P)\) denote the affine hull of \(P\), namely
the smallest affine subspace containing \(P\).
If \(P=\conv\{u_1,\ldots,u_s\}\), then \(A\) consists of all affine
combinations \(\sum_{i=1}^s\lambda_i u_i\), where
\(\lambda_i\in\R\) for \(i\in[s]\) and
\(\sum_{i=1}^s\lambda_i=1\); the coefficients may be negative.
Define
\[
L=\{u-v:u,v\in A\}.
\]
The set \(L\) is a linear subspace of \(\R^n\), called the direction
space of \(A\). For any fixed \(a_0\in A\),
\[
A=\{a_0+\ell:\ell\in L\},
\]
so \(A\) is a translate of the linear space \(L\).
The dimension of \(P\) is defined as the dimension of its affine hull;
thus,
\[
\dim P=\dim L.
\]
Interiors and boundaries of polytopes are understood relative to their
affine hulls.

For \(a\in L\setminus\{0\}\), let \(\alpha=\max_{x\in P}a\cdot x\).
The hyperplane \(\{x\in A:a\cdot x=\alpha\}\) is a supporting
hyperplane of \(P\), and its intersection with \(P\),
\[
F=\{x\in P:a\cdot x=\alpha\},
\]
is the face exposed by \(a\), with \(a\) as an outward normal.
Every nonempty proper face of a convex polytope admits such a
representation. The sets \(\varnothing\) and \(P\) are also regarded
as faces. When \(d=\dim P\ge1\), faces of dimensions \(0\), \(1\), and
\(d-1\) are called vertices, edges, and facets,
respectively. In particular, the facets of a three-dimensional polytope
are two-dimensional faces.

Every edge of a three-dimensional convex polytope is contained in exactly
two facets and equals their intersection. If these facets are exposed by
outward normals \(a_1,a_2\), then \(a_1+a_2\) exposes their common edge:
equality in the sum of the two supporting inequalities holds exactly
when equality holds in both. Moreover, the vertex--edge graph of a convex
polytope is connected. Any two facets of a three-dimensional convex
polytope can also be joined by a sequence of facets in which consecutive
facets share an edge.

For the cube \(Q=[-1,1]^m\), every nonempty face is obtained by fixing
some coordinates to \(+1\) or \(-1\). Specifically, for \(I\subseteq[m]\)
and \(\varepsilon_i\in\{-1,1\}\), the set
\[
\{z\in Q:z_i=\varepsilon_i\text{ for all }i\in I\}
\]
is a face of dimension \(m-|I|\). For \(z\in \partial Q\), the smallest face
containing \(z\) fixes precisely the coordinates satisfying \(|z_i|=1\);
all remaining coordinates vary freely in \((-1,1)\).

\subsection{Homotopy, contractibility, and the degree of sphere maps}\label{subsec:topology-preliminaries}

The upper-bound proof compares sphere maps through homotopies and derives
a contradiction from their degrees. We recall the required concepts and
properties~\citep{hatcher2002}.

Let \(X,Y\) be topological spaces. A homeomorphism is a continuous
bijection with a continuous inverse. Two continuous maps \(f,g:X\to Y\)
are homotopic, written \(f\simeq g\), if there exists a continuous
map \(H:X\times[0,1]\to Y\) such that, for every \(x\in X\),
\[
H(x,0)=f(x),\qquad H(x,1)=g(x).
\]
The map \(H\) is called a homotopy from \(f\) to \(g\).

A nonempty space \(X\) is contractible if its identity map is
homotopic to a constant map. Both the constant map and the homotopy
take values in \(X\). For example, a nonempty convex set is
contractible: fixing \(x_0\in X\), the map \(H(x,t)=(1-t)x+tx_0\)
gives a homotopy from the identity to the constant map \(x\mapsto x_0\).

For an integer \(d\ge1\), write
\(S^d=\{x\in\R^{d+1}:\|x\|=1\}\).
Its \(d\)-th singular homology group with integer coefficients satisfies
\(H_d(S^d;\mathbb Z)\cong\mathbb Z\).
A continuous map \(f:S^d\to S^d\) induces a homomorphism on this group,
necessarily given by multiplication by an integer. This integer is the
degree of \(f\), denoted by \(\deg f\). We use the following
properties:
\begin{enumerate}
\item If \(f\simeq g\), then \(\deg f=\deg g\).
\item The identity satisfies \(\deg(\operatorname{id}_{S^d})=1\).
\item If \(f(S^d)\ne S^d\), then \(\deg f=0\). In particular, a constant
map has degree zero.
\end{enumerate}

The third property can be understood through contractibility.
If \(f\) omits a point \(p\in S^d\), its image lies in
\(S^d\setminus\{p\}\). This space is homeomorphic to \(\R^d\) and is
therefore contractible, so \(f\) is homotopic to a constant map.

In applications, sphere-valued homotopies can be constructed by
normalizing nonzero vectors. If
\(B:X\times[0,1]\to\R^{d+1}\setminus\{0\}\) is continuous, then
\[
\widehat B(x,t)=\frac{B(x,t)}{\|B(x,t)\|}
\]
is a sphere-valued homotopy. Thus, when using linear interpolation to
construct such a homotopy, the interpolating vectors must remain
nonzero throughout.

Finally, for \(m\ge2\), the boundary of the cube \(Q=[-1,1]^m\) is
homeomorphic to \(S^{m-1}\) through the radial map
\[
\mathcal R:\partial Q\to S^{m-1},
\qquad
\mathcal R(z)=\frac{z}{\|z\|},
\]
whose inverse is
\[
\mathcal R^{-1}(y)=\frac{y}{\max_{1\le j\le m}|y_j|}.
\]
Consequently, if \(F:Q\to\R^m\) is continuous and \(F(z)\ne0\)
for every \(z\in\partial Q\), we can define
\[
\widehat F(z)=\frac{F(z)}{\|F(z)\|},
\qquad z\in\partial Q.
\]
Composing \(\widehat F\) with
\(\mathcal R^{-1}:S^{m-1}\to\partial Q\) gives the sphere self-map
\(f:S^{m-1}\to S^{m-1}\), defined by
\(f(y)=\widehat F(\mathcal R^{-1}(y))\). We can therefore consider
the degree of \(f\).

\subsection{Main result}

We now state the main result for the hypothesis class \(\mathcal H_\Psi\).

\begin{theorem}[Exact dimension]\label{thm:main}
For the class \(\mathcal H_\Psi\) of predictors with pairwise distinct
class weights in Definition~\ref{def:model}, if
\(n\ge3\) and \(k\ge3\), then
\begin{equation}\label{eq:main}
\Ndim(\mathcal H_\Psi)=n(k-1)-1.
\end{equation}
\end{theorem}
Theorem~\ref{thm:main} follows from Propositions~\ref{prop:lower}
and~\ref{prop:upper}: Proposition~\ref{prop:lower} in Section~\ref{sec:lower} constructs a shattered
set of this size, and Proposition~\ref{prop:upper} in Section~\ref{sec:upper} proves that no larger
set is shattered.

\section{A lower bound}\label{sec:lower}

To establish the lower bound, we construct a set of \(n(k-1)-1\)
feature vectors together with two distinct candidate labels for each
vector. We then show that every choice of one candidate label for each
vector can be realized by a suitable choice of pairwise distinct class
weights.

\begin{proposition}[Lower bound]\label{prop:lower}
For \(n,k\ge3\), we have
\begin{equation}\label{eq:lower}
\Ndim(\mathcal H_\Psi)\ge n(k-1)-1.
\end{equation}
\end{proposition}

The idea is to control the score difference within each candidate pair
while keeping both candidate scores above those of all other labels.
The proof proceeds in three steps. In Step~1, we choose class weights
as vertices of a polytope and use its exposed edges to select feature
vectors and fix their candidate label pairs. In Step~2, we form the
comparison matrix \(\mathcal T\), show that its rank is \(n(k-1)-1\),
and select \(n(k-1)-1\) independent rows to obtain a submatrix
\(T\) of full row rank. The corresponding feature vectors form the
set to be shattered. In Step~3, we use a right inverse of
\(T\) to construct weight perturbations realizing every candidate
labeling. A single sufficiently small perturbation scale preserves
the strict score advantage of both candidate labels over the other
labels and keeps the class weights pairwise distinct.

\begin{proof}
\emph{Step 1: exposed edges and candidate label pairs.}
We first choose class weight vectors \(u_i\in\R^n\), \(i\in[k]\).
For \(k=3\), take \(u_1=0\), \(u_2=e_1\), and \(u_3=e_2\), where
\(e_1,e_2\) are the first two standard basis vectors of \(\R^n\).
Their convex hull is a nondegenerate triangle. For \(k\ge4\), choose
distinct real numbers \(t_1,\ldots,t_k\) and set
\begin{equation}\label{eq:moment-curve}
u_i=(t_i,t_i^2,t_i^3,0,\ldots,0),
\qquad P=\conv\{u_1,\ldots,u_k\}.
\end{equation}
Each \(u_i\) is a vertex: the functional with coefficient vector
\((2t_i,-1,0,\ldots,0)\) takes value
\(t_i^2-(t_j-t_i)^2\) at \(u_j\), uniquely maximizing at \(j=i\).
Any four vertices are affinely independent, since the matrix with rows
\((1,t_i,t_i^2,t_i^3)\) has a nonzero Vandermonde determinant.
Consequently \(P\) is three-dimensional and each of its facets is a
triangle. When \(n>3\), these dimensions refer to its affine hull.

Let \(E\) be the edge set of \(P\), or the three edges of the triangle
when \(k=3\), identifying each edge \([u_i,u_j]\) with its endpoint
index pair \(\{i,j\}\). For each \(\{i,j\}\in E\), define
\begin{equation}\label{eq:edge-inputs}
\begin{split}
H_{ij}&=\{x\in\R^n:(u_j-u_i)\cdot x=0\},\\
\Omega_{ij}&=
\{x\in H_{ij}:(u_i-u_d)\cdot x>0
                   \text{ for all }d\in[k]\setminus\{i,j\}\}.
\end{split}
\end{equation}
Fix an edge \(\{i,j\}\in E\). By the exposed-face property recalled in
Section~\ref{subsec:polytope-preliminaries}, there exists \(x_0\in\R^n\)
such that the linear functional \(u\mapsto u\cdot x_0\) attains its
maximum over \(\conv\{u_1,\ldots,u_k\}\) exactly on \([u_i,u_j]\).
Thus the two endpoint weights have the same score at \(x_0\), strictly
greater than that of every other class. Hence \(x_0\in\Omega_{ij}\),
so \(\Omega_{ij}\) is nonempty. By definition, for every
\(x\in\Omega_{ij}\) and every \(d\in[k]\setminus\{i,j\}\),
\begin{equation}\label{eq:exposed-edge}
u_i\cdot x=u_j\cdot x>u_d\cdot x.
\end{equation}

Since \(u_i\ne u_j\), the space \(H_{ij}\) has dimension \(n-1\).
The set \(\Omega_{ij}\) is defined within \(H_{ij}\) by finitely many
strict linear inequalities, so it is open in \(H_{ij}\). Therefore,
there exists \(\eta>0\) such that
\[
\{x\in H_{ij}:\|x-x_0\|<\eta\}\subseteq\Omega_{ij}.
\]
Differences of points in this ball span \(H_{ij}\). These differences
belong to \(\Span(\Omega_{ij})\), and hence
\(\Span(\Omega_{ij})=H_{ij}\).

For each edge \(\{i,j\}\in E\), choose \(n-1\) linearly independent
feature vectors in \(\Omega_{ij}\), and assign the candidate labels
\(i\) and \(j\) to each of them.
All selected feature vectors are distinct. Within each edge, this
follows from linear independence. Vectors selected for different edges
cannot coincide, since the pair of maximizing class labels uniquely
determines the edge.

\emph{Step 2: the rank of the edge comparisons.}
We next determine how many of these candidate comparisons can be varied
independently.
We form the comparison matrix \(\mathcal T\) from all feature vectors
chosen in Step~1, using the candidate pair assigned to each vector.
The matrix has \(|E|(n-1)\) rows.
For each edge \(\{i,j\}\in E\) with \(i<j\), and each of its
\(n-1\) selected feature vectors \(x\), the corresponding row is
indexed by \((ij,x)\) and has \(k\) blocks of size \(1\times n\).
For \(c\in[k]\), its \(c\)-th block is defined by
\[
\mathcal T_{(ij,x),c}=
\begin{cases}
-x^\top,&c=i,\\
 x^\top,&c=j,\\
 0_{1\times n},&c\notin\{i,j\}.
\end{cases}
\]
Thus, for \(Z=(z_1,\ldots,z_k)\in\R^{nk}\),
\[
(\mathcal T Z)_{ij,x}=(z_j-z_i)\cdot x.
\]
We claim that
\begin{equation}\label{eq:kernel}
\ker\mathcal T=
\{(b+su_1,\ldots,b+su_k):b\in\R^n,\ s\in\R\}.
\end{equation}

Fix \(Z\in\ker\mathcal T\). For each edge \(\{i,j\}\in E\), the feature
vectors chosen from \(\Omega_{ij}\) span \(H_{ij}\). Hence, for every
edge \(\{i,j\}\in E\) with \(i<j\),
there exists \(\lambda_{ij}\in\R\) such that
\begin{equation}\label{eq:edge-multiplier}
z_j-z_i=\lambda_{ij}(u_j-u_i).
\end{equation}
On a triangular face with vertices \(u_i,u_j,u_\ell\), where
\(i<j<\ell\), put
\(a=u_j-u_i\) and \(b'=u_\ell-u_i\). The identity
\((z_j-z_i)+(z_\ell-z_j)-(z_\ell-z_i)=0\) yields
\[
(\lambda_{ij}-\lambda_{j\ell})a
+(\lambda_{j\ell}-\lambda_{i\ell})b'=0.
\]
Since \(a,b'\) are independent, the three edge multipliers agree.
For \(k=3\), this covers every edge. For \(k\ge4\), triangular
facets are connected through shared edges: radial projection from an
interior point identifies the boundary with a triangulated sphere, and
paths between facet interiors can avoid vertices and cross successive
edges. Equality therefore propagates to every edge. Let \(s\) be the
common multiplier. The vertex-edge graph is connected as well, so
\[
z_i-z_1=s(u_i-u_1),\qquad z_i=b+su_i,
\quad b=z_1-su_1.
\]
Conversely, every tuple of this form lies in \(\ker\mathcal T\),
because \((u_j-u_i)\cdot x=0\) for \(x\in H_{ij}\).
The parameterization by \((b,s)\) is injective: if every
\(b+su_i=0\), subtracting equations for two distinct vertices gives
\(s=0\), and then \(b=0\). This proves~\eqref{eq:kernel} and gives
\begin{equation}\label{eq:rank}
\dim\ker\mathcal T=n+1,
\qquad \rank\mathcal T=nk-(n+1)=n(k-1)-1.
\end{equation}

Since \(\rank\mathcal T=n(k-1)-1\), we can select
\(m:=n(k-1)-1\) linearly independent rows of \(\mathcal T\),
obtaining a full-row-rank matrix
\(T\in\R^{m\times nk}\). Denote the corresponding distinct inputs
by \(x_1,\ldots,x_m\) and their fixed candidate pairs by
\(a_i<b_i\).
Thus, for any block vector \(Z=(z_1,\ldots,z_k)\), with
\(z_c\in\R^n\), we have
\[
(TZ)_i=(z_{b_i}-z_{a_i})\cdot x_i,\qquad i\in[m].
\]

\emph{Step 3: a uniform perturbation for all candidate labelings.}
It remains to realize every choice between the selected candidates
without changing their dominance over the other labels.
Take the right inverse of \(T\)
\begin{equation}\label{eq:right-inverse}
R=T^\top(TT^\top)^{-1},\qquad TR=I_m.
\end{equation}
For each \(\sigma\in\sgnset^m\), write
\(R\sigma=(\delta_1,\ldots,\delta_k)\), where
\(\delta_c=(R\sigma)_c\in\R^n\) is the \(c\)-th vector block
for each \(c\in[k]\). These blocks depend on \(\sigma\).
Define the following uniform bounds over all choices:
\begin{equation}\label{eq:uniform-constants}
\begin{aligned}
D&=\max_{\sigma\in\sgnset^m,\ c\in[k]}\|(R\sigma)_c\|>0,\\
X&=\max_{1\le i\le m}\|x_i\|>0.
\end{aligned}
\end{equation}
Define the base margin and weight separation explicitly by
\[
\begin{aligned}
g&=\min_{\substack{1\le i\le m,\ c\in\{a_i,b_i\}\\
                     d\in[k]\setminus\{a_i,b_i\}}}
       (u_c-u_d)\cdot x_i>0,\\
\rho&=\min_{1\le c<d\le k}\|u_c-u_d\|>0.
\end{aligned}
\]
Both minima are taken over finite nonempty sets. Their positivity
follows from the strict score inequalities above and the pairwise
distinctness of \(u_1,\ldots,u_k\). Choose one
\(\varepsilon>0\), independently of \(\sigma\), such that
\begin{equation}\label{eq:epsilon}
2\varepsilon DX<g,\qquad 2\varepsilon D<\rho.
\end{equation}
With this \(\varepsilon\) fixed, take an arbitrary
\(\sigma\in\sgnset^m\) and set \(w_c=u_c+\varepsilon\delta_c\)
for every \(c\in[k]\), using the blocks of \(R\sigma\).
For every \(i\in[m]\), the identity
\(u_{a_i}\cdot x_i=u_{b_i}\cdot x_i\), together with
\(TR\sigma=\sigma\), gives
\begin{equation}\label{eq:target-signs}
(w_{b_i}-w_{a_i})\cdot x_i=\varepsilon\sigma_i.
\end{equation}
For every \(i\in[m]\), every \(c\in\{a_i,b_i\}\), and every
\(d\in[k]\setminus\{a_i,b_i\}\), Cauchy--Schwarz yields
\begin{equation}\label{eq:external-margin}
(w_c-w_d)\cdot x_i
\ge g-\varepsilon\|\delta_c-\delta_d\|\,\|x_i\|
\ge g-2\varepsilon DX>0.
\end{equation}
For every \(1\le c<d\le k\), the weights satisfy
\begin{equation}\label{eq:distinct-margin}
\|w_c-w_d\|\ge\rho-2\varepsilon D>0.
\end{equation}
For \(\sigma_i=-1\) the unique winner is \(a_i\); for
\(\sigma_i=+1\) it is \(b_i\). The inputs and candidate pairs
were fixed before \(\sigma\) was chosen, and the same
\(\varepsilon\) works for all \(2^m\) choices.
This proves~\eqref{eq:lower} with strict winning margins.
\end{proof}

\begin{example}[An explicit five-point construction in \(\R^3\)]\label{ex:five}
Take \(u_1=0\), \(u_2=e_1\), and \(u_3=e_2\).
The exposed-edge regions are
\(\Omega_{12}=\{(0,y,z):y<0\}\),
\(\Omega_{13}=\{(x,0,z):x<0\}\), and
\(\Omega_{23}=\{(t,t,z):t>0\}\).
The following five distinct inputs lie in these regions:
\begin{center}
\begin{tabular}{ccc}
\toprule
\(i\)&\(x_i\)&\(\{a_i,b_i\}\)\\
\midrule
1 & \((0,-1,0)\) & \(\{1,2\}\)\\
2 & \((0,-1,1)\) & \(\{1,2\}\)\\
3 & \((-1,0,0)\) & \(\{1,3\}\)\\
4 & \((-1,0,1)\) & \(\{1,3\}\)\\
5 & \((1,1,0)\) & \(\{2,3\}\)\\
\bottomrule
\end{tabular}
\end{center}
For any \(\sigma\in\sgnset^5\), set \(\varepsilon=1/20\) and
\[
\begin{aligned}
w_1&=0,\\
w_2&=e_1+\varepsilon
\begin{pmatrix}\sigma_1-\sigma_3-\sigma_5\\-\sigma_1\\\sigma_2-\sigma_1\end{pmatrix},\\
w_3&=e_2+\varepsilon
\begin{pmatrix}-\sigma_3\\0\\\sigma_4-\sigma_3\end{pmatrix}.
\end{aligned}
\]
Direct substitution gives
\((w_{b_i}-w_{a_i})\cdot x_i=\varepsilon\sigma_i\) for all
five points. The perturbation blocks have norms at most
\(D=\sqrt{14}\); the base quantities satisfy
\(g=\rho=1\) and \(X=\sqrt2\). Hence
\(2\varepsilon DX=\sqrt{28}/10<1\) and
\(2\varepsilon D=\sqrt{14}/10<1\).
Equations~\eqref{eq:external-margin}--\eqref{eq:distinct-margin}
therefore ensure unique winners and pairwise distinct weights for
all \(32\) candidate labelings. Thus the five-point set is shattered
by \(\mathcal H_\Psi\), giving
\(\Ndim(\mathcal H_\Psi)\ge5\).
\end{example}

\section{An upper bound}\label{sec:upper}

In this section, we prove the upper bound.
We first establish comparison independence and strict avoidance of the
candidate labels, then use induction and cube interpolation to
derive a contradiction by computing the degree of the associated
sphere map in two ways.
Fix \(k\ge3\). We use \(V_r\) from~\eqref{eq:Vr}.

For the induction below, an indexed candidate system
\((x_i,a_i,b_i)_{i=1}^m\) records feature vectors \(x_i\) and fixed
candidate labels \(a_i<b_i\) in \([k]\); the feature vectors may
coincide. Realizability of every candidate labeling by \(\mathcal H_\Psi\)
means that, for every tuple \((y_1,\ldots,y_m)\) with
\(y_i\in\{a_i,b_i\}\) for all \(i\in[m]\), there exists
\(h\in\mathcal H_\Psi\) satisfying \(h(x_i)=y_i\) for all \(i\in[m]\).
After projection, we retain each index and its candidate pair and
apply the same realization condition to the projected feature vectors.
Keeping the indices allows the induction to proceed without a
general-position assumption.

\subsection{Comparison independence and strict avoidance}

The following lemma adapts the linear-dependence argument for
homogeneous halfspaces in the proof of Theorem~9.2 of
\citet{shalevshwartz2014} to the effective parameter space \(V_r\),
with the smallest-label tie rule made explicit.

\begin{lemma}\label{lem:independence}
Let \((x_i,a_i,b_i)_{i=1}^m\) be an indexed candidate system in
\(\R^r\), with \(a_i<b_i\).
For \(v=(v_2,\ldots,v_k)\in V_r\), let \(h_v\) denote the prediction
rule associated with the full tuple \((0,v_2,\ldots,v_k)\).
Suppose that, for every
\((y_i)_{i=1}^m\in\{a_1,b_1\}\times\cdots\times\{a_m,b_m\}\), there exists
\(v\in V_r\) such that \(h_v(x_i)=y_i\) for all \(i\in[m]\),
under the smallest-label tie rule. Then the functionals
\begin{equation}\label{eq:comparisons}
\rho_i(v)=(v_{b_i}-v_{a_i})\cdot x_i,
\qquad 1\le i\le m,
\end{equation}
are linearly independent on \(V_r\). Consequently \(m\le qr\).
If \(m=qr\), the map
\begin{equation}\label{eq:upper-T}
T:V_r\longrightarrow\R^m,
\qquad T(v)=(\rho_1(v),\ldots,\rho_m(v))
\end{equation}
is a linear isomorphism. For each fixed candidate pair, the indexed
inputs carrying that pair are linearly independent.
\end{lemma}

\begin{proof}
Choosing \(a_i\) requires \(\rho_i(v)\le0\), whereas choosing
\(b_i\) requires \(\rho_i(v)>0\), since a tie would favor
\(a_i<b_i\). Suppose \(\sum_i t_i\rho_i=0\) is a nontrivial
relation. Reverse its sign if necessary so that some \(t_i>0\).
Choose \(b_i\) when \(t_i>0\), choose \(a_i\) when \(t_i<0\),
and choose either candidate when \(t_i=0\). At a realizing parameter,
every \(t_i\rho_i(v)\) is nonnegative and at least one is strictly
positive, contradicting the relation. A dependence among inputs with a
common candidate pair would induce the same dependence among their
functionals. The dimension
bound and the isomorphism statement follow. In particular, under the
hypotheses of the lemma, every \(x_i\) is nonzero.
\end{proof}

It shows that realizing every candidate labeling forces the comparison
functionals to be linearly independent, thereby giving the upper bound
\(n(k-1)\) on the Natarajan dimension, which is consistent with the
bound established in \citet{pan2026}.

The following lemma, together with interpolation, will be used
in Step~3 of the proof of Proposition~\ref{prop:upper} below.

\begin{lemma}[Strict avoidance]\label{lem:avoidance}
Let \(k\ge3\), and let \((x_i,a_i,b_i)_{i=1}^m\) be an indexed
candidate system in \(\R^r\), with \(m\ge1\). Suppose that, for every
\((y_1,\ldots,y_m)\in\{a_1,b_1\}\times\cdots\times\{a_m,b_m\}\),
there exists \(v\in\mathcal A_r\), corresponding to pairwise distinct
weights, such that \(h_v(x_i)=y_i\) for all \(i\in[m]\), under the
smallest-label tie rule. Then there exists
a single \(u\in V_r\setminus\{0\}\) such that, for every \(i\in[m]\),
the maximum class score at \(x_i\) under \(u\) is attained by
exactly one label, which lies outside \(\{a_i,b_i\}\).
\end{lemma}

\begin{proof}
\emph{Case \(k=3\).}
Let \(A,B,C\) denote the inputs with candidate pairs \(\{1,2\}\),
\(\{1,3\}\), and \(\{2,3\}\), respectively. Choose a weight tuple
\(W=(w_1,\ldots,w_k)\) realizing the labeling
\[
A\mapsto2,\qquad B\mapsto1,\qquad C\mapsto3.
\]
By Lemma~\ref{lem:independence}, the inputs in \(B\) are linearly
independent. Hence there is \(d\in\R^r\) such that
\(d\cdot x=1\) for every \(x\in B\); take \(d=0\) if \(B\) is
empty. Replace \(w_1\) by \(w_1+\varepsilon d\). On \(B\), every
margin of label 1 increases by \(\varepsilon\), making label 1 a
unique winner. On \(A\) and \(C\), the desired winner already
strictly beats label 1. Choose one positive \(\varepsilon\) small
enough to preserve these strict inequalities simultaneously at all
inputs in \(A\cup C\). Comparisons between
labels 2 and 3 are unchanged.

Subtract the new first weight from all weights. Labels 1 on \(B\) and
3 on \(C\) are now unique winners. On \(A\), label 2 strictly beats
label 1 but may tie with label 3 at a common positive score \(h\).
Choose one \(0<\eta<1/2\), small enough to preserve all existing
strict winning comparisons at all inputs, and replace
\[
w_2\mapsto(1-\eta)w_2,
\qquad w_3\mapsto(1-2\eta)w_3.
\]
All existing strict margins remain positive, and a tie on \(A\)
becomes a margin \(\eta h>0\) in favor of label 2. Thus, at every
feature vector, the prescribed label uniquely maximizes the class score.
Apply the permutation
\(1\mapsto2\mapsto3\mapsto1\) to the weights. The winners on
\(A,B,C\) become \(3,2,1\), respectively, and lie outside their
candidate pairs.

\emph{Case \(k\ge4\).}
Let \(\pi\) fix label 1 and cycle the others:
\[
2\mapsto3\mapsto\cdots\mapsto k\mapsto2.
\]
For each \(i\in[m]\), choose \(c_i\in\{a_i,b_i\}\setminus\{1\}\)
such that
\(\pi(c_i)\notin\{a_i,b_i\}\). Such an endpoint exists. If the
pair contains 1, choose its other endpoint. Otherwise, if both images
remained in the pair, that two-element set would be invariant under a
cycle of length \(k-1\ge3\), which is impossible.

Choose a weight tuple \(W=(w_1,\ldots,w_k)\) realizing the labels
\(c_i\), and subtract the first weight from every class weight so that
\(w_1=0\).
Since \(c_i>1\), its winning score \(w_{c_i}\cdot x_i\) is
strictly positive. Choose one \(0<\varepsilon<1/k\), small enough to preserve all
strict winning comparisons simultaneously for every input, and set
\[
\widetilde w_j=(1-\varepsilon j)w_j,
\qquad 1\le j\le k.
\]
Every previously strict winning comparison remains strict. If \(c_i\)
originally tied with \(j\ne c_i\), then \(c_i<j\), and their
common score \(h\) is positive. The new difference is
\[
(1-\varepsilon c_i)h-(1-\varepsilon j)h
=\varepsilon(j-c_i)h>0.
\]
Thus every chosen label becomes a unique winner. Relabel the weights by
\(\widetilde w^{\mathrm{new}}_{\pi(j)}=\widetilde w_j\). Their
winners are \(\pi(c_i)\), outside the prescribed pairs.

In both cases, take the effective parameter of the final tuple to
obtain \(u\). It is nonzero because the system is nonempty and its
winners are strict. The perturbations can be chosen small enough that
the weights remain pairwise distinct. Relabeling is performed only
after the winners have become
unique, so no permutation invariance of the tie rule is assumed.
\end{proof}

\subsection{Induction and cube interpolation}

We now establish the upper bound for \(\mathcal H_\Psi\).

\begin{proposition}[Upper bound]\label{prop:upper}
For \(n,k\ge3\), we have
\begin{equation}\label{eq:upper}
\Ndim(\mathcal H_\Psi)\le n(k-1)-1.
\end{equation}
\end{proposition}

Lemma~\ref{lem:independence} gives \(m\le qr\) and makes the
comparison map \(T\) a linear isomorphism when \(m=qr\).
Thus, it remains to exclude equality. The proof combines induction
on the dimension of the feature space with two computations of the
degree of a sphere self-map.

In Step~1, we set up the induction and interpolate the parameters
realizing the candidate labelings over a cube \(Q\). Applying
the comparison map \(T\) gives a map \(F\) satisfying sign conditions
on the cube faces. In Step~2, we use projection and the induction
hypothesis to show that \(F\) does not vanish on \(\partial Q\),
so its restriction to the boundary can be normalized. In Step~3,
Lemma~\ref{lem:avoidance} and convex interpolation show that
\(F(\partial Q)\) avoids a positive ray. The associated sphere
self-map therefore omits a point and has degree zero. In Step~4,
a small translation makes the face signs strict while the map remains
nonzero on the boundary throughout the translation. We then show that
the associated sphere self-map is homotopic to the identity, giving
degree one and yielding the required contradiction.

\begin{proof}
For each integer \(r\ge0\), let \(P(r)\) assert that, for every
indexed candidate system with \(qr\) inputs in \(\R^r\), there
exists at least one candidate labeling that no parameter in
\(\mathcal A_r\) realizes.
We prove \(P(r)\) by induction for all \(r\ge0\), including
the lower-dimensional systems produced by projection.

\emph{Step 1: the induction and the interpolating map.}
For \(r=0\), the space \(\R^0\) has only one vector, so no tuple
of \(k\) pairwise distinct weights exists. Even the empty input list
has one candidate labeling, which would require such a tuple. Hence
\(P(0)\) holds.

Assume \(r\ge1\) and \(P(t)\) for every integer \(0\le t<r\),
and suppose \(P(r)\) fails. Fix an indexed candidate system
\((x_i,a_i,b_i)_{i=1}^m\), with \(m=qr\ge2\), for which every
candidate labeling is realized by a parameter in \(\mathcal A_r\).
Lemma~\ref{lem:independence} makes
\(T:V_r\to\R^m\) in~\eqref{eq:upper-T} an isomorphism, and all
inputs are nonzero. For each \(s\in\sgnset^m\), choose one
realization whose first class weight is zero,
with effective parameter \(v_s\in\mathcal A_r\),
where \(s_i=-1\)
selects \(a_i\) and \(s_i=+1\) selects \(b_i\). Each chosen tuple
has pairwise distinct weights.
Write \(v_s=(v_{s,2},\ldots,v_{s,k})\), where
\(v_{s,c}\in\R^r\) for \(2\le c\le k\), and set \(v_{s,1}=0\).

Set \(Q=[-1,1]^m\). For every \(z\in Q\) and
\(s\in\sgnset^m\), define
\begin{equation}\label{eq:cube-weights}
\lambda_s(z)=\prod_{i=1}^m\frac{1+s_i z_i}{2},
\end{equation}
\begin{equation}\label{eq:interpolation}
G(z)=\sum_{s\in\sgnset^m}\lambda_s(z)v_s,
\qquad F(z)=T(G(z)).
\end{equation}
For every \(z\in Q\), the coefficients are nonnegative and sum
to one. Write \(G(z)=(G_2(z),\ldots,G_k(z))\), where
\(G_c(z)\in\R^r\) for \(2\le c\le k\), and set \(G_1(z)=0\).
Write \(F(z)=(F_1(z),\ldots,F_m(z))\). By the definition of \(T\),
for every \(z\in Q\) and \(i\in[m]\), we have
\[
F_i(z)=\rho_i(G(z))
=\bigl(G_{b_i}(z)-G_{a_i}(z)\bigr)\cdot x_i.
\]

A vertex
\(s\) is said to contribute at \(z\) if \(\lambda_s(z)>0\).
For each \(i\in[m]\), only vertices selecting \(a_i\) contribute
on the face \(z_i=-1\), and only vertices selecting \(b_i\)
contribute on the face \(z_i=+1\). Thus, for every \(i\in[m]\)
and every \(z\in Q\),
\begin{equation}\label{eq:face-signs}
z_i=-1\ \Longrightarrow\ F_i(z)\le0,
\qquad z_i=+1\ \Longrightarrow\ F_i(z)>0.
\end{equation}
The interpolated weights need not be pairwise distinct. Interpolation
takes place in the full linear space; admissibility is required for the
vertex realizations and for the lower-dimensional system used below.

\emph{Step 2: exclusion of boundary zeros.}
We next show that the map \(F\) constructed in Step~1 is nonzero on
\(\partial Q\). A boundary zero would yield a lower-dimensional
candidate system that contradicts the induction hypothesis. Excluding
such zeros will allow us to normalize the boundary restriction of
\(F\) for the degree argument.
Suppose \(F(z)=0\) for some \(z\in\partial Q\). Invertibility of
\(T\) gives \(G(z)=0\). Let
\[
I=\{i\in[m]:|z_i|=1\},\qquad J=[m]\setminus I.
\]
The set \(I\) is nonempty. On the smallest face containing \(z\),
the coordinates indexed by \(I\) are fixed and those indexed by \(J\)
are free. More explicitly, a vertex contributes precisely when
\(s_i=z_i\) for every \(i\in I\). For each such vertex,
\[
\lambda_s(z)=\prod_{j\in J}\frac{1+s_jz_j}{2}>0,
\]
since \(|z_j|<1\) on \(J\); all other coefficients vanish.
Thus the contributing vertices realize every candidate choice on
\(J\), while the choices on \(I\) remain fixed.

By~\eqref{eq:face-signs}, \(z_i=-1\) for every \(i\in I\).
Hence every contributing vertex realization predicts \(a_i\) at \(x_i\)
for each \(i\in I\). If \(a_i>1\) for some \(i\in I\),
every contributing
vertex would give that label a strictly positive score relative to
label 1, whose score is zero. The convex average of these positive
scores would also be positive, contradicting \(G(z)=0\). Thus
\(a_i=1\) for every \(i\in I\).
For every \(i\in I\), every \(c\in\{2,\ldots,k\}\), and every
\(s\in\sgnset^m\) with \(\lambda_s(z)>0\),
\[
v_{s,c}\cdot x_i\le0.
\]
The weighted average of these nonpositive values, with positive
coefficients \(\lambda_s(z)\) for the contributing vertices, is
\(G_c(z)\cdot x_i=0\), so every value is zero. Set
\begin{equation}\label{eq:UL}
U=\Span\{x_i:i\in I\},\qquad
\ell=\dim U\ge1,\qquad L=(U^\perp)^q.
\end{equation}
Here \((U^\perp)^q\) denotes the Cartesian product of \(q=k-1\)
copies of \(U^\perp\); explicitly,
\[
\begin{split}
L=\{(v_2,\ldots,v_k):{}&v_c\in U^\perp\\
&\text{for every }c=2,\ldots,k\}.
\end{split}
\]
The notation \(U^q\) analogously denotes the Cartesian product of
\(q\) copies of \(U\).
All contributing parameters belong to \(L\), whose dimension is
\(q(r-\ell)\). If \(r-\ell=0\), every class weight of a
contributing vertex is zero, contradicting pairwise distinctness.
Hence \(r-\ell\ge1\).

Project the inputs indexed by \(J\) orthogonally onto \(U^\perp\)
and use orthonormal coordinates on that space. For each contributing
vertex \(s\), all class weights in the tuple \((0,v_{s,2},\ldots,v_{s,k})\)
lie in \(U^\perp\). Thus projection preserves every score and predicted
label. For every contributing vertex \(s\) and every
\(1\le c<d\le k\), the difference \(v_{s,c}-v_{s,d}\) is nonzero
in \(U^\perp\), and the coordinate isomorphism preserves this property.
The new tuples therefore still have pairwise distinct weights.
For every candidate labeling on \(J\), a contributing vertex provides
a realization after projection. Applying Lemma~\ref{lem:independence}
to these projected feature vectors in \(\R^{r-\ell}\) gives
\[
|J|\le q(r-\ell).
\]
Identifying the comparison functionals with their Euclidean coefficient
vectors, those indexed by \(I\) lie in \(U^q\) and are independent.
Indeed, \(a_i=1\) makes their only nonzero coefficient block
equal to \(x_i\) in block \(b_i\). Since \(\dim U^q=q\ell\),
we have \(|I|\le q\ell\). Since \(|I|+|J|=qr\), both bounds are
equalities:
\begin{equation}\label{eq:IJ-count}
|I|=q\ell,\qquad |J|=q(r-\ell).
\end{equation}

By~\eqref{eq:IJ-count}, the projected tuples realize every candidate
labeling of \(q(r-\ell)\) indexed feature vectors in dimension
\(r-\ell<r\), contradicting \(P(r-\ell)\). Consequently
\begin{equation}\label{eq:boundary-nonzero}
\forall z\in\partial Q,\qquad F(z)\ne0.
\end{equation}

\emph{Step 3: an omitted direction and degree zero.}
Step~2 ensures that \(F\) can be normalized on \(\partial Q\). We next
combine Lemma~\ref{lem:avoidance} with convex interpolation to show
that its boundary image avoids a positive ray. The resulting sphere
map will then omit a point and have degree zero.

Fix the single parameter \(u\) supplied by
Lemma~\ref{lem:avoidance}. Take any \(z\in\partial Q\) and choose
an index \(i\in[m]\) with \(|z_i|=1\), which exists because \(z\)
is a boundary point. This index may depend on \(z\). Set \(c=a_i\)
if \(z_i=-1\), and \(c=b_i\) if \(z_i=+1\). Every vertex
contributing at \(z\) has this same winner \(c\) at \(x_i\).
Hence, for every such vertex \(s\) and every \(d\in[k]\),
\[
(v_{s,c}-v_{s,d})\cdot x_i\ge0.
\]
Convex averaging shows that label \(c\) attains the maximum score
at \(x_i\) under the interpolated weights \(G(z)\), possibly tied
with other labels. For every \(t>0\), however, \(tu\) has a unique
winner outside \(\{a_i,b_i\}\), so \(G(z)\ne tu\).
Since \(T\) is a linear isomorphism and \(z\) was arbitrary,
we obtain, for every \(z\in\partial Q\) and every \(t>0\),
\begin{equation}\label{eq:ray-avoidance}
F(z)\ne tT(u).
\end{equation}

Let \(S^{m-1}\) be the Euclidean unit sphere and let
\[
\mathcal R:\partial Q\longrightarrow S^{m-1},
\qquad \mathcal R(z)=z/\|z\|
\]
be the radial homeomorphism. By~\eqref{eq:boundary-nonzero}, the
restriction of \(F:Q\to\R^m\) to \(\partial Q\) can be normalized
to give
\[
\widehat F:\partial Q\longrightarrow S^{m-1},
\qquad \widehat F(z)=\frac{F(z)}{\|F(z)\|}.
\]
To obtain a sphere self-map whose degree we can compute, we use
\(\mathcal R^{-1}:S^{m-1}\to\partial Q\) to define
\[
\begin{aligned}
f:S^{m-1}&\longrightarrow S^{m-1},\\
f(y)&=\widehat F(\mathcal R^{-1}(y))\\
&=\frac{F(\mathcal R^{-1}(y))}
        {\|F(\mathcal R^{-1}(y))\|}.
\end{aligned}
\]
Since \(m\ge2\),
the sphere has positive dimension. We use the standard facts
that degree is invariant under homotopy, that the identity has degree
one, and that a map omitting a point has degree zero because the
punctured sphere is contractible~\citep{hatcher2002}.
Since \(u\ne0\) and \(T\) is invertible, \(T(u)\ne0\).
Equation~\eqref{eq:ray-avoidance} implies that \(f\) omits
\(T(u)/\|T(u)\|\), so
\begin{equation}\label{eq:degree-zero}
\deg f=0.
\end{equation}

\emph{Step 4: a homotopy to the identity and degree one.}
To obtain a contradiction with~\eqref{eq:degree-zero}, we return to
the face conditions in~\eqref{eq:face-signs}. A small translation of
\(F\) will make these signs strict while keeping the map nonzero on
the boundary. This will allow us to construct a homotopy from
\(f\) to the identity and show that \(\deg f=1\).

First, compactness and~\eqref{eq:boundary-nonzero} give
\[
\beta=\min_{z\in\partial Q}\|F(z)\|>0.
\]
Continuity and~\eqref{eq:face-signs} also give
\[
\alpha=\min_{1\le i\le m}\ \min_{\substack{z\in Q\\z_i=1}}F_i(z)>0.
\]
Each inner minimum is over a compact face, and there are finitely
many such faces. Write \(\one=(1,\ldots,1)\in\R^m\).
Choose a single \(\delta>0\) satisfying the bound below and define
\(F_\delta:Q\to\R^m\) by
\begin{equation}\label{eq:shift}
0<\delta<\min\left\{\alpha,\frac{\beta}{2\sqrt m}\right\},
\qquad F_\delta(z)=F(z)-\delta\one.
\end{equation}
For \(0\le t\le1\) and \(z\in\partial Q\),
\[
\|F(z)-t\delta\one\|
\ge\beta-\delta\sqrt m>0.
\]
Define \(\widehat F_\delta:\partial Q\to S^{m-1}\) by
\[
\widehat F_\delta(z)=\frac{F_\delta(z)}{\|F_\delta(z)\|},
\qquad z\in\partial Q.
\]
Normalization therefore provides a homotopy between \(\widehat F\)
and \(\widehat F_\delta\). Moreover, for every \(i\in[m]\) and every \(z\in Q\),
\[
z_i=-1\Longrightarrow(F_\delta)_i(z)<0,
\qquad
z_i=+1\Longrightarrow(F_\delta)_i(z)>0.
\]
For every \(t\in[0,1]\) and \(z\in\partial Q\), define
\[
H_t(z)=(1-t)F_\delta(z)+tz.
\]
Choose an index \(i\in[m]\) with \(|z_i|=1\). The two
\(i\)-th coordinates in this convex combination both have the strict
sign of \(z_i\), so \((H_t(z))_i\ne0\).
Thus \(H_t(z)\ne0\) for every \(t\in[0,1]\) and every
\(z\in\partial Q\).
After normalization, this gives a homotopy
from \(\widehat F_\delta\) to \(\mathcal R\). The small
translation uses the boundary nonvanishing already proved in Step~2;
no strictification of the original vertex realizations is assumed.
Precomposing both homotopies with \(\mathcal R^{-1}\) gives
\[
\begin{aligned}
f=\widehat F\circ\mathcal R^{-1}
&\simeq\widehat F_\delta\circ\mathcal R^{-1}\\
&\simeq\mathcal R\circ\mathcal R^{-1}
=\operatorname{id}_{S^{m-1}}.
\end{aligned}
\]
Hence \(\deg f=1\), contradicting~\eqref{eq:degree-zero}.
\end{proof}

\begin{example}[continued]
For \(n=k=3\), the effective parameter space
\(V_3=\R^6\) has dimension \(6\).
By Lemma~\ref{lem:independence}, any shattered set has at most six points.
The induction and cube-interpolation argument in
Proposition~\ref{prop:upper} excludes the six-point case, so
\[
\Ndim(\mathcal H_\Psi)\le3(3-1)-1=5.
\]
The five-point construction in Example~\ref{ex:five} attains this bound.
Consequently,
\[
\Ndim(\mathcal H_\Psi)=5.
\]
\end{example}

\section{Conclusion}\label{sec:conclusion}

For homogeneous multiclass linear predictors with pairwise distinct
class weights, we prove \(\Ndim(\mathcal H_\Psi)=n(k-1)-1\) for
\(n,k\ge3\) by establishing matching
lower and upper bounds.
The present result, together with the results in \citet{shalevshwartz2014}
and \citet{pan2026}, determines the Natarajan dimension of this hypothesis
class for all \(n,k\ge2\).


\begin{thebibliography}{12}

\bibitem[Ben-David et~al.(1995)Ben-David, Cesa-Bianchi, Haussler, and Long]{bendavid1995}
Ben-David, S., Cesa-Bianchi, N., Haussler, D., Long, P. M., 1995.
\newblock Characterizations of learnability for classes of \(\{0,\ldots,n\}\)-valued functions.
\newblock \emph{Journal of Computer and System Sciences} \textbf{50} (1), 74--86.
\newblock DOI: \href{https://doi.org/10.1006/jcss.1995.1008}{10.1006/jcss.1995.1008}.

\bibitem[Blumer et~al.(1989)Blumer, Ehrenfeucht, Haussler, and Warmuth]{blumer1989}
Blumer, A., Ehrenfeucht, A., Haussler, D., Warmuth, M. K., 1989.
\newblock Learnability and the Vapnik--Chervonenkis dimension.
\newblock \emph{Journal of the ACM} \textbf{36} (4), 929--965.
\newblock DOI: \href{https://doi.org/10.1145/76359.76371}{10.1145/76359.76371}.

\bibitem[Daniely et~al.(2012)Daniely, Sabato, and Shalev-Shwartz]{daniely2012}
Daniely, A., Sabato, S., Shalev-Shwartz, S., 2012.
\newblock Multiclass learning approaches: A theoretical comparison with implications,
in: Pereira, F., Burges, C. J. C., Bottou, L., Weinberger, K. Q. (Eds.),
\emph{Advances in Neural Information Processing Systems}, vol. 25.
\newblock Curran Associates, Inc., pp. 485--493.
\newblock URL: \url{https://proceedings.neurips.cc/paper_files/paper/2012/hash/19f3cd308f1455b3fa09a282e0d496f4-Abstract.html}.

\bibitem[Daniely and Shalev-Shwartz(2014)]{daniely2014}
Daniely, A., Shalev-Shwartz, S., 2014.
\newblock Optimal learners for multiclass problems,
in: Balcan, M. F., Feldman, V., Szepesv\'{a}ri, C. (Eds.),
\emph{Proceedings of the 27th Conference on Learning Theory}, Barcelona, Spain.
\newblock Proceedings of Machine Learning Research, vol. 35, PMLR, pp. 287--316.
\newblock URL: \url{https://proceedings.mlr.press/v35/daniely14b.html}.

\bibitem[Daniely et~al.(2015)Daniely, Sabato, Ben-David, and Shalev-Shwartz]{daniely2015}
Daniely, A., Sabato, S., Ben-David, S., Shalev-Shwartz, S., 2015.
\newblock Multiclass learnability and the ERM principle.
\newblock \emph{Journal of Machine Learning Research} \textbf{16} (72), 2377--2404.
\newblock URL: \url{https://www.jmlr.org/papers/v16/daniely15a.html}.

\bibitem[Hatcher(2002)]{hatcher2002}
Hatcher, A., 2002.
\newblock \emph{Algebraic Topology}.
\newblock Cambridge University Press, Cambridge.
\newblock URL: \url{https://pi.math.cornell.edu/~hatcher/AT/ATpage.html}.

\bibitem[Jin(2023)]{jin2023}
Jin, Y., 2023.
\newblock Upper bounds on the Natarajan dimensions of some function classes,
in: \emph{2023 IEEE International Symposium on Information Theory (ISIT)}, Taipei, Taiwan.
\newblock IEEE, pp. 1020--1025.
\newblock DOI: \href{https://doi.org/10.1109/ISIT54713.2023.10206618}{10.1109/ISIT54713.2023.10206618}.

\bibitem[Natarajan(1989)]{natarajan1989}
Natarajan, B. K., 1989.
\newblock On learning sets and functions.
\newblock \emph{Machine Learning} \textbf{4} (1), 67--97.
\newblock DOI: \href{https://doi.org/10.1007/BF00114804}{10.1007/BF00114804}.

\bibitem[Pan and Li(2026)]{pan2026}
Pan, Y., Li, B., 2026.
\newblock Bounds on the Natarajan dimension of a class of linear multi-class predictors.
\newblock \emph{Pattern Recognition Letters} \textbf{200}, 129--134.
\newblock DOI: \href{https://doi.org/10.1016/j.patrec.2025.12.012}{10.1016/j.patrec.2025.12.012}.

\bibitem[Shalev-Shwartz and Ben-David(2014)]{shalevshwartz2014}
Shalev-Shwartz, S., Ben-David, S., 2014.
\newblock \emph{Understanding Machine Learning: From Theory to Algorithms}.
\newblock Cambridge University Press, Cambridge.
\newblock DOI: \href{https://doi.org/10.1017/CBO9781107298019}{10.1017/CBO9781107298019}.

\bibitem[Vapnik and Chervonenkis(1971)]{vapnik1971}
Vapnik, V. N., Chervonenkis, A. Ya., 1971.
\newblock On the uniform convergence of relative frequencies of events to their probabilities.
\newblock \emph{Theory of Probability \& Its Applications} \textbf{16} (2), 264--280.
\newblock DOI: \href{https://doi.org/10.1137/1116025}{10.1137/1116025}.

\bibitem[Ziegler(1995)]{ziegler1995}
Ziegler, G. M., 1995.
\newblock \emph{Lectures on Polytopes}.
\newblock Graduate Texts in Mathematics, vol. 152. Springer, New York.
\newblock DOI: \href{https://doi.org/10.1007/978-1-4613-8431-1}{10.1007/978-1-4613-8431-1}.

\end{thebibliography}
\end{document}